\documentclass[10pt,twocolumn,letterpaper]{article}
\usepackage{cvpr}   
\usepackage[utf8]{inputenc}
\usepackage{graphicx}
\usepackage{amsmath}
\usepackage{amssymb}
\usepackage{booktabs}

\usepackage[utf8]{inputenc}
\usepackage{epsfig}
\usepackage{amsmath}
\usepackage{amssymb}
\usepackage{listings}
\usepackage{times}
\usepackage{epsfig}
\usepackage{graphicx}
\usepackage{stfloats}
\usepackage{amsmath}
\usepackage{amssymb}
\usepackage{capt-of}
\usepackage{graphicx}
\usepackage{multirow}
\usepackage{stmaryrd}
\usepackage[noend]{algpseudocode}
\usepackage{float}
\usepackage{booktabs,xcolor}
\definecolor{darkgreen}{rgb}{0.0, 0.2, 0.13}
\usepackage[pagebackref,breaklinks,colorlinks]{hyperref}

\usepackage[capitalize]{cleveref}
\crefname{section}{Sec.}{Secs.}
\Crefname{section}{Section}{Sections}
\Crefname{table}{Table}{Tables}
\crefname{table}{Tab.}{Tabs.}

\begin{document}
\title{Learning a Weight Map for Weakly-Supervised Localization}

\author{Tal Shaharabany\\
Tel-Aviv University\\
{\tt\small shaharabany@mail.tau.ac.il}
\and
Lior Wolf\\
Tel-Aviv University\\
{\tt\small liorwolf@gmail.com}
}

\maketitle

\begin{abstract}
In the weakly supervised localization setting, supervision is given as an image-level label. We propose to employ an image classifier $f$ and to train a generative network $g$ that outputs, given the input image, a per-pixel weight map that indicates the location of the object within the image. Network $g$ is trained by minimizing the discrepancy between the output of the classifier $f$ on the original image and its output given the same image weighted by the output of $g$. The scheme requires a regularization term that ensures that $g$ does not provide a uniform weight, and an early stopping criterion in order to prevent $g$ from over-segmenting the image. Our results indicate that the method outperforms existing localization methods by a sizable margin on the challenging fine-grained classification datasets, as well as a generic image recognition dataset. Additionally, the obtained weight map is also state-of-the-art in weakly supervised segmentation in fine-grained categorization datasets.
\end{abstract}

\section{Introduction}

While deep learning is often said to require large amounts of labeled data, the study of neural networks brought a major revolution in unsupervised and weakly supervised learning. For example, the advent of GANs ~\cite{goodfellow2014generative} led to an unprecedented ability to generate images, powerful unsupervised techniques now exist for mapping samples across multiple domains e.g.,\cite{zhu2017unpaired,choi2018stargan}, and unsupervised image representation methods learn powerful representations without labeled data e.g., \cite{oord2018representation,he2019momentum}. 

The accuracy of supervised image localization methods has improved dramatically in the last decade \cite{ren2015faster, liu2016ssd, duan2019centernet}. However, these methods rely on bounding-box supervision, which is not always available. Weakly supervised methods have emerged as an alternative that relies only on image-level labeling to one of the multiple classes. 

Most weakly supervised localization methods rely on the assumption that a trained image classifier $f$ relies on image regions that are within the foreground segment and try to analyze the behavior of the classifier to extract this information \cite{zhou2016learning, qin2019rethinking}. Breaking down this assumption one can identify three challenges: first, the classifier can rely on context, i.e., on regions outside the object. Second, the classifier may rely on parts of the object and ignore much of it. Third, there is the problem of the explainability of the classifier; i.e., building a causal model that interprets their behavior is an open problem.

\begin{figure}[t]
    \begin{center}
    \begin{tabular}{@{}c@{~~}c@{~~}c@{~~}c@{~~}c@{}}
    \multirow{0}{*}{\raisebox{-.755in}[0pt][0pt]{\includegraphics[scale=2]{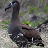}}}&
    \includegraphics[width=0.125\linewidth]{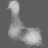}&
    \includegraphics[width=0.125\linewidth]{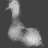}&
    \includegraphics[width=0.125\linewidth]{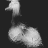}&
    \includegraphics[width=0.125\linewidth]{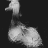}\\
    &\includegraphics[width=0.125\linewidth]{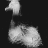}&
    \includegraphics[width=0.125\linewidth]{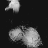}&
    \includegraphics[width=0.125\linewidth]{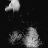}&
    \includegraphics[width=0.125\linewidth]{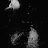}\\
    &\includegraphics[width=0.125\linewidth]{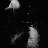}&
    \includegraphics[width=0.125\linewidth]{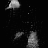}&
    \includegraphics[width=0.125\linewidth]{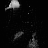}&
    \includegraphics[width=0.125\linewidth]{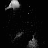} \\
    \end{tabular}
    \end{center}
\caption{Given an input image (left), our method trains a generative network to output weights such that a weighted image would be classified similarly to the input image. As training progresses (small panels), the generated weights tend to become more specific. An automatic stopping criterion is used to return the checkpoint in which the weight map provides good localization.} 
    \label{fig:teaser}
\end{figure}

For the first problem, one can attempt to rely on explainability methods that differentiate between positive and negative contributions ~\cite{nam2019relative, gur2021visualization}, or assume, as we do, that modern classifiers are less prone to such issues, especially when delineating between similar classes, which tend to appear in similar contexts. 

The second challenge, which may lead to the identification of parts instead of the entire object, is a major issue in modern weakly supervised localization methods ~\cite{zhang2018adversarial, zhang2018self, xue2019danet}. In our case, we offer to solve it by an early stopping technique, since the generative method we propose evolves to become increasingly localized during the training process. Our method, in contrast to many weakly supervised approaches, employs a segmentation-like network $g$ that provides a pixel-wise weight map given the input image ($I$). Since it is trained on all images, rather than solving one image at a time, it learns general patterns before it learns to identify specific image locations associated with a given class. 

An example is shown in Fig.~\ref{fig:teaser}, demonstrating an image from the CUB-200-2011 bird dataset. The output of $g$ is shown for consecutive epochs. As can be seen, the outline of the bird is first identified. As training progresses, the network learns to output specific parts of the bird, which are most relevant to distinguishing it from other birds. It is also evident that the transition between these two modes can be found by considering the completeness of the obtained shape. Averaged over many images, this provides a robust stopping criterion.

\begin{figure*}[t]
    \begin{center}
    \begin{tabular}{c}
    \includegraphics[width=1\linewidth]{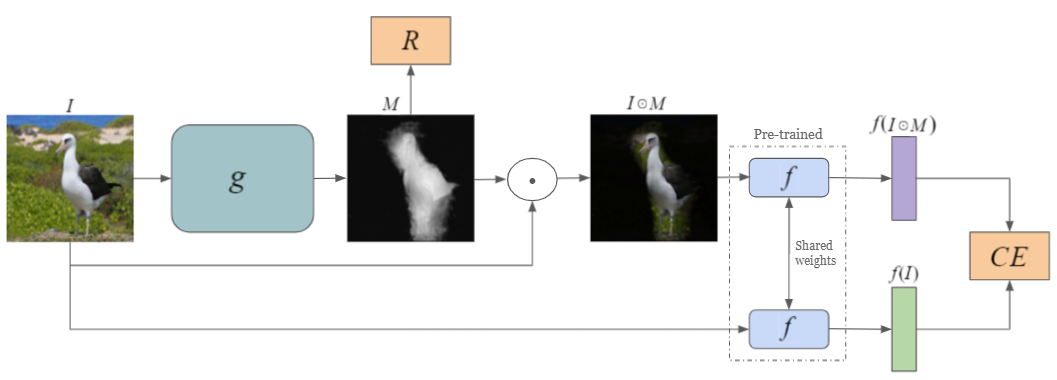} 
    \end{tabular}
    \end{center}
    \caption{The proposed method employs, similarly to other methods, an image-level pre-trained classifier $f$. It generates a per-pixel weight map for the input image $I$, using a learned network $g$. This network is optimized so that the pseudo-probabilities $f$ outputs on $I$ are close (in terms of CE=the Cross-Entropy loss) to the ones it outputs on weighted image $I\cdot g(i)$. A regularization loss term, denoted by $R$, encourages $M=g(I)$ to be sparse.}
    \label{fig:arch}
\end{figure*}

The third challenge we have pointed to was the challenge of extracting the relevant information from the classifier $f$. While many methods rely on the activations of the network and on its gradients e.g., ~\cite{sundararajan2017axiomatic,smilkov2017smoothgrad,selvaraju2017grad,bach2015pixel}, this is known to be an unreliable process~\cite{asano2019critical}. In fact, one can already point to the collinearity problem in linear regression as presenting ambiguity regarding the causal links between input and prediction~\cite{dormann2013collinearity}.

Our approach, instead of trying to analyze the behavior of the classifier $f$ given $I$ by considering the information flow within $f$, derives a training loss for training network $g$ that only considers the output of $f$. Specifically, we require that the pseudo probabilities provided by $f$ given the input image be the same as the vector of pseudo probabilities $f$ outputs outputs on the image $I\cdot g(I)$, which is a pixel-wise weighting of an image by the weight mask produced by $g$.

{In another form of our method, we replace the classifier by a Siamese network. Two types of triplet losses are then used to train network. The first one tries to separate between the foreground the background of the anchor image, while keeping the representation of the entire image close to that of the foreground.  The second loss aims to keep the latent distance close of masked foreground images of the same class, and distance the latent representation of foreground images from different classes.} 

Our experiments, similarly to other recent contributions, focus on datasets for fine-grained classification, in which the problem of weakly supervised localization is more challenging, due to the need of $f$ to attend to well-localized differences between the classes. As our results show, using the label-difference criteria on a pretrained classifier or Siamese network, in order to train a segmentation network $g$, leads to results that are far superior to the state-of-the-art methods.

\section{Related Work}
Supervised deep learning algorithms often require a training set that relies on extensive human annotation.  This is especially the case for segmentation \cite{long2015fully, zhao2018icnet} and object detection \cite{ren2015faster, liu2016ssd, zhou2019objects}.

In order to decrease the dependency on human annotation, weakly supervised approaches have been developed \cite{bilen2016weakly, jie2017deep}. In this paper, we focus on image-level annotation in order to predict the localization of the object bounding box. 

Fine-grained recognition datasets have become extremely popular as benchmarks for weakly supervised object localization (WSOL). In fine-grained recognition, the annotation is more challenging, since it requires professional knowledge and domain experts, for instance, of Ornithology for the CUB-200-2011 dataset \cite{wah2011caltech}. Since fine-grained recognition methods require specialization in order to differentiate between similar classes, and since this specialization is often accompanied by the extraction of localized features~\cite{lin2015bilinear,yu2018hierarchical,gao2016compact,fu2017look,peng2017object, yang2018learning}, the information extracted from such classifiers is often less accessible when trying to segment the entire object. 

Many algorithms were proposed for the task of WSOL. The Class Activation Map (CAM) explainability method \cite{zhou2016learning} and its variants \cite{qin2019rethinking} identify the salient pixels that lead to the classification. A multi-task loss function proposed by \cite{lu2020geometry} takes shape into consideration. Adversarial Complementary Learning (ACoL) \cite{zhang2018adversarial} employs two parallel classifiers where complementary regions are discovered via an adversarial erasing of feature maps. The divergent activation approach (DA) \cite{xue2019danet}  aggregates and shares information from different spatial layers of the backbone. A similarity score that contrasts high-activation regions with other regions was proposed by \cite{zhang2020rethinking}. 

A common improvement is to add a localization assistance branch to the CNN classifier \cite{zhang2018self, zhang2020inter}. While this creates clear separation between the supervised classification goal and the indirectly supervised localization goal, it only partly solves the challenge of focusing on the most discriminative regions at the expense of detecting the entire object. Other attempts to solve this problem mask random patches in the training set, thus forcing the network to rely less on well-localized cues \cite{singh2017hide, bazzani2016self, yun2019cutmix}. Similarly, an attention-based dropout layer encourages the network to also consider less discriminative parts of the image \cite{choe2019attention}. 

Previous WSOL works have been criticized by \cite{choe2020evaluating} for selecting the best checkpoint and the hyperparameters by considering the test data. It also offers an evaluation metric for weakly supervised segmentation where instead of bounding-box annotation, a foreground-background mask is given. Unlike conventional segmentation metrics, one considers multiple thresholds, since tuning the threshold is challenging in the weakly supervised setting. Very recently, \cite{choe2021region} presented a new method for weighting the feature maps. Evaluation is done following the stringent settings presented by \cite{choe2020evaluating}, and weakly-supervised segmentation results are presented on Oxford-flower102 \cite{nilsback2008automated}.

Our work mitigates the challenge imposed by the locality of discriminative features by employing early stopping. Our intuition is that the learned localization network becomes increasingly specific as training progresses. The usage of early stopping before the network is fully trained for a given task is reminiscent of the usage of early stopping in the Deep image prior technique of \cite{ulyanov2018deep}, where it is used to prevent a network that is trained to reconstruct the input image from overfitting on it, thus allowing it to express patterns that match the inductive bias of CNNs.

\begin{figure}[t]
    \begin{center}
    \begin{tabular}{cc}
    \includegraphics[width=0.895\linewidth]{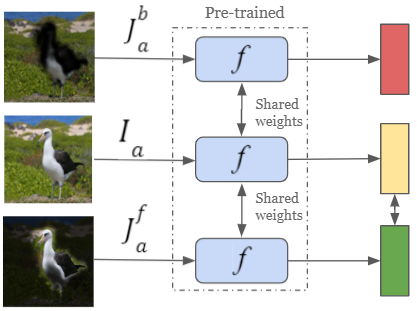} \\ (a) \\
    \includegraphics[width=0.895\linewidth]{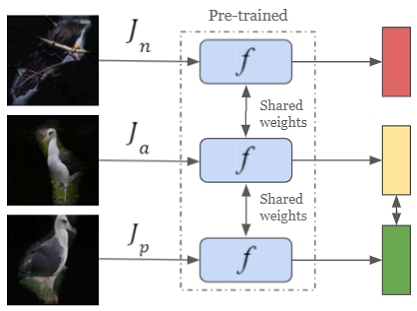} \\ (b) \\
    \end{tabular}
    \end{center}
    \caption{Two triplet losses are used for training $g$  for method II, where $f$ is a pretrained supervised Siamese network. (a) The inner loss brings together the image representation (yellow box) and the masked image representation (green box), while distancing the background representation (red box) from the original image representation. (b) The outer loss is a triplet loss that distances the foreground representation of same class images (anchor and positive samples), while distancing that of different classes (anchor and negative samples).
    }
    \label{fig:triplet}
\end{figure}

\section{Method}
This section introduces the proposed methods for weakly supervised fined-grained object localization. One method is based on employing a classifier $f$ that receives an image and outputs a pseudo probability vector. The other method employs a Siamese network that combines two copies of $f$ with shared weights. The two methods provide similar results, and throughout the paper we present sample outputs for the first method. See supplementary for visual results of the second method.

Both methods employ an encoder-decoder network $g$ that maps the input image $I \in  R^{3 \times W \times H}$ to pixel-wise weight mask $M\in  R^{W \times H}$. Such networks are typically used in supervision segmentation \cite{long2015fully, ronneberger2015u} and object detection \cite{ren2015faster} tasks, where the ground truth map is given and the network weights are updated accordingly. Under the weakly supervised object localization (WSOL) settings, the ground truth is not given locally (per pixel). Instead, only a global label of the image is given. 

 \begin{figure*}[t]
    \setlength{\tabcolsep}{4.5pt} 
    \renewcommand{\arraystretch}{2} 
    \begin{tabular}{ccccccc}
    Epoch 1 & Epoch 4 & Epoch 7 & Epoch 10 & Epoch 13 & Epoch 22 & Epoch 31 \\
    \vspace{-.4cm}
    \includegraphics[width=0.12\linewidth]{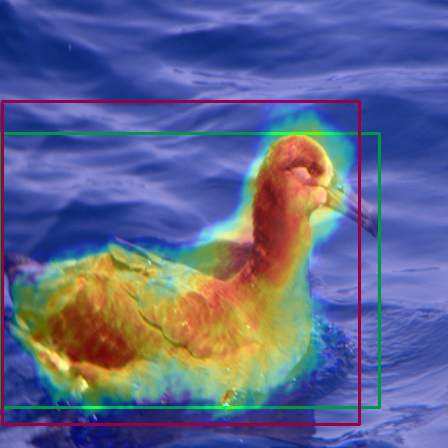} &
    \includegraphics[width=0.12\linewidth]{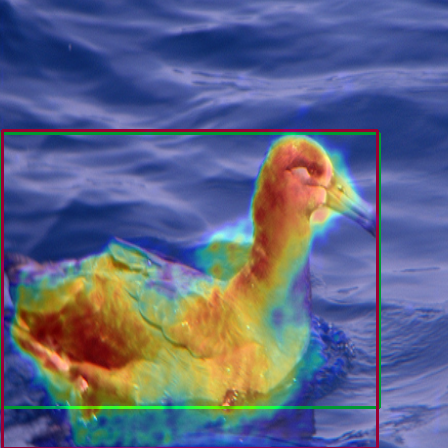} &
    \includegraphics[width=0.12\linewidth]{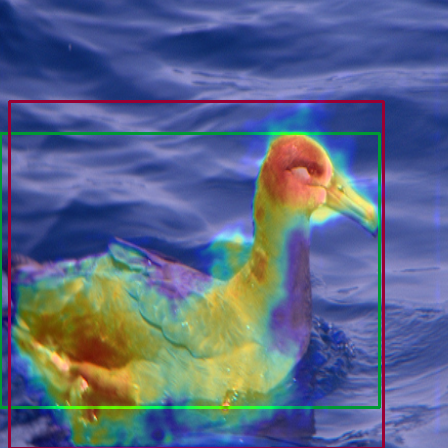} &
    \includegraphics[width=0.12\linewidth]{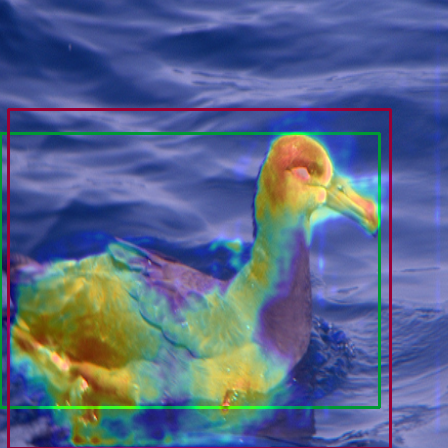} &
    \includegraphics[width=0.12\linewidth]{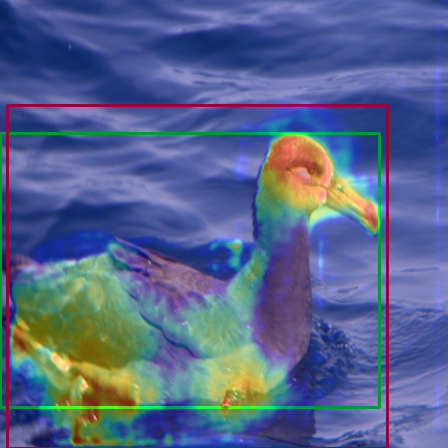} &
    \includegraphics[width=0.12\linewidth]{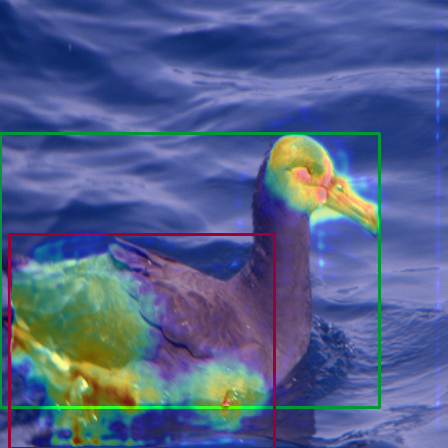} &
    \includegraphics[width=0.12\linewidth]{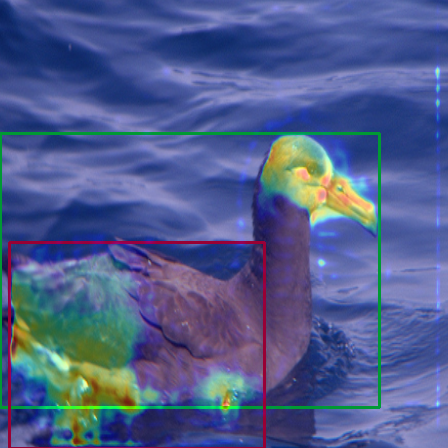} \\
    (IOU = 0.8067) & (IOU = 0.8543) & (IOU = 0.7698) & (IOU = 0.7751) & (IOU = 0.7738) & (IOU = 0.4021) & (IOU = 0.3705) \\
    \vspace{-.4cm}
    \includegraphics[width=0.12\linewidth]{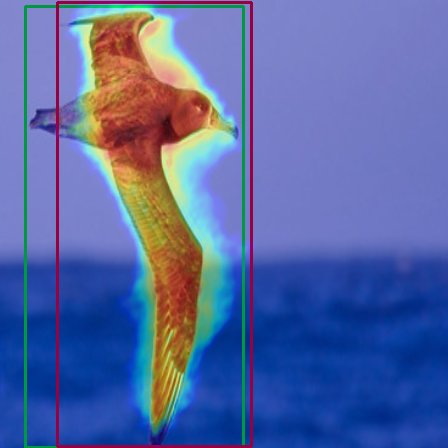} &
    \includegraphics[width=0.12\linewidth]{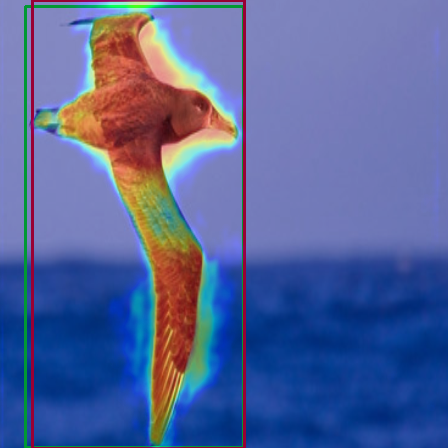} &
    \includegraphics[width=0.12\linewidth]{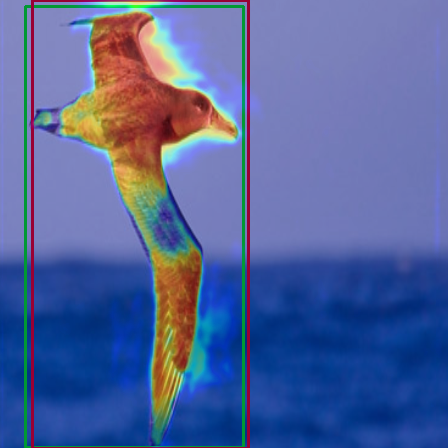} &
    \includegraphics[width=0.12\linewidth]{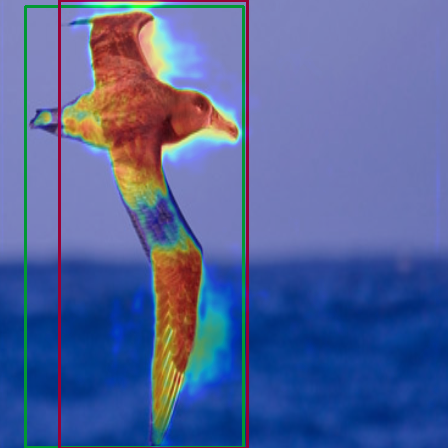} &
    \includegraphics[width=0.12\linewidth]{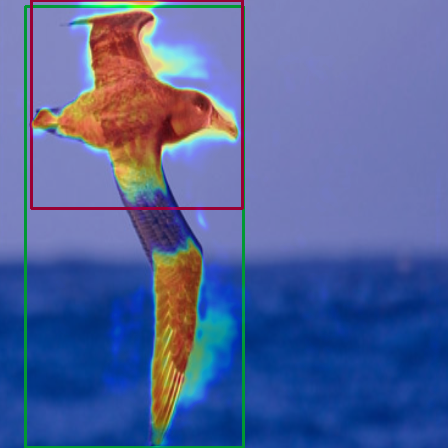} &
    \includegraphics[width=0.12\linewidth]{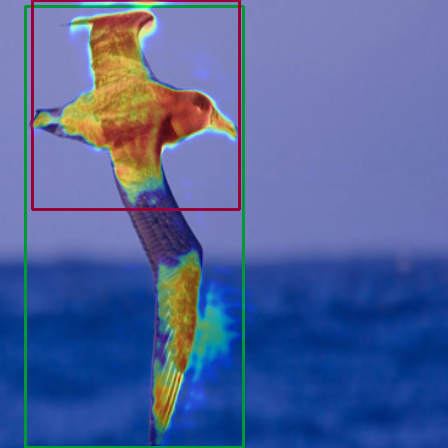} &
    \includegraphics[width=0.12\linewidth]{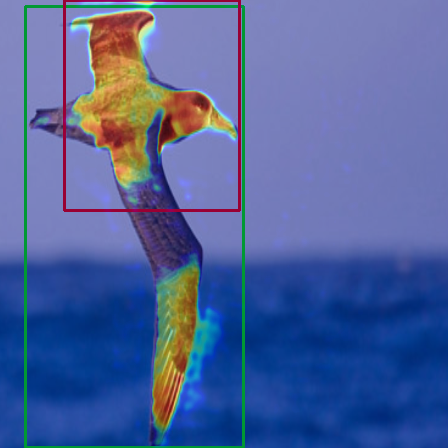} \\
    (IOU = 0.8182) & (IOU = 0.9517) & (IOU = 0.9347) & (IOU = 0.8212) & (IOU = 0.4375) & (IOU = 0.4315) & (IOU = 0.3677) \\
    \vspace{-.4cm}
    \includegraphics[width=0.12\linewidth]{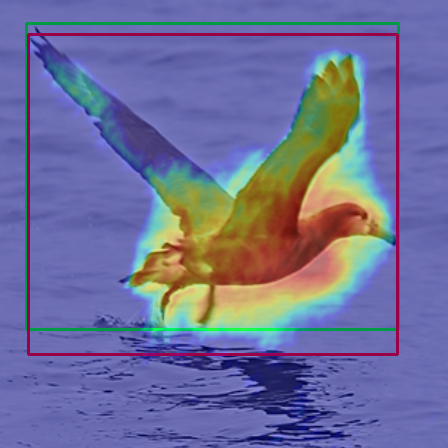} &
    \includegraphics[width=0.12\linewidth]{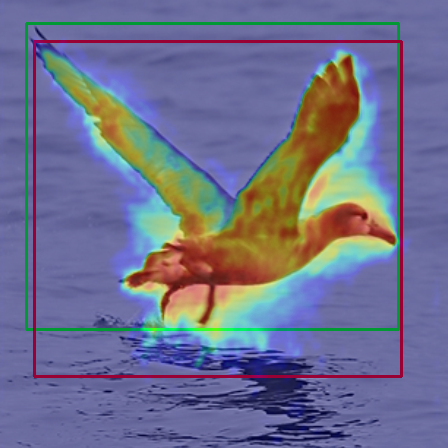} &
    \includegraphics[width=0.12\linewidth]{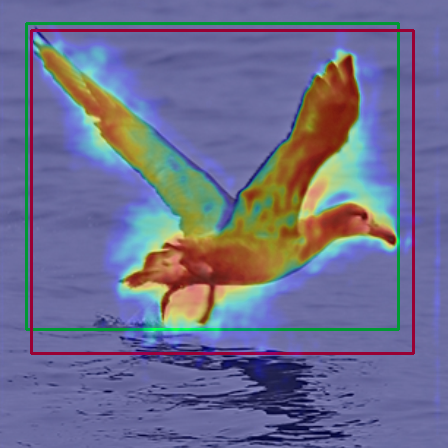} &
    \includegraphics[width=0.12\linewidth]{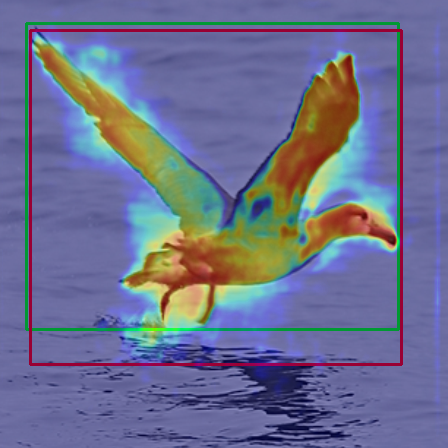} &
    \includegraphics[width=0.12\linewidth]{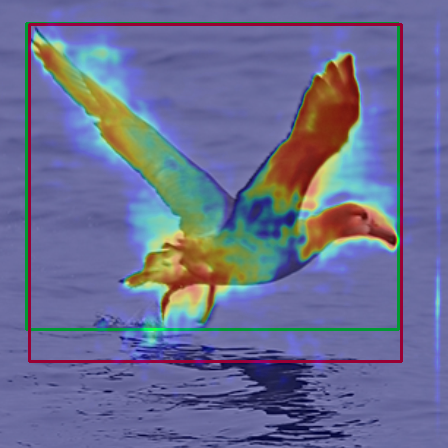} &
    \includegraphics[width=0.12\linewidth]{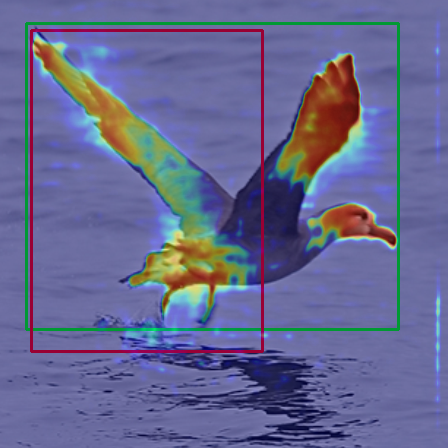} &
    \includegraphics[width=0.12\linewidth]{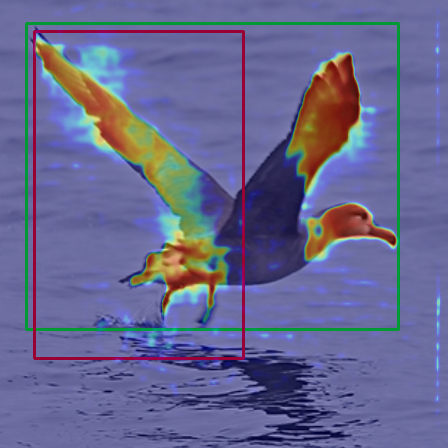} \\
    (IOU = 0.8884) & (IOU = 0.7983) & (IOU = 0.8647) & (IOU = 0.8655) & (IOU = 0.8927) & (IOU = 0.5839) & (IOU = 0.5225) \\
    
    \end{tabular}
    \caption{The weight map generated by network $g$ for various epochs on the CUB dataset. Also shown are the obtained bounding boxes. Our stopping criterion relies on some of the higher weights being left out of the obtained bounding boxes when $g$ starts to over-segment.} 
    \label{fig:EarlyStop}
\end{figure*}

\paragraph{Methods I (classifier based)}
To train  network $g$ under WSOL settings, we offer a novel two-stage algorithm: (1) train a classifier  $f: R^{3 \times W \times H} \to R^{d}$, with the global supervision, where d is the number of categories in the dataset, and (2) freeze the classifier weights and employ a classification-invariance loss function for the training of $g$:
\begin{equation}
\label{eq:one}
   \min_{ g } D(f(I),f(I \odot g(I))) + \lambda R(g(I)),
\end{equation}
where D measures the discrepancy between the classifier output, $R$ is a regularizer that minimizes the size of the highlighted region produced by the weight mask $M=g(I)$, and $\lambda$ is a weight factor. The goal of this loss is to maintain the same classifier representation $f$ with and without masking by $g$, see Fig~\ref{fig:arch}. In our implementation, the cross-entropy loss is used for $D$, and $R$ is the L1 norm. For the sake of simplicity, $\lambda$ is set to one and no tuning of this value was ever performed.  

 Without the regularization term $R(g(I))$, the architecture would converge to a naive solution where the output mask is fixed at one. With the regularization term, the network is encouraged to provide more localized solutions, in which the non-relevant attributes are ignored.

\paragraph{Methods II (Siamese network)}

The second method follows the architectures and algorithms given by Method I. However, instead of a classifier $f$, it compares an anchor image $I_a$ to a same class (``positive'') image $I_p$ vs. a negative image from a different class $I_n$. In the case of the 2nd method, $f$ is pretrained using the following triplet loss:
\begin{equation}
    \label{eq:triplet}
 L =  \max \big\{  \|f(I_a) - f(I_p) \|  - \|f(I_a) - f(I_n) \| + 1 , 0  \big\}
 \end{equation}

Once $f$ is trained, we train network $g$ to minimize two triplet loss terms:
\begin{equation}
\label{eq:triplet1}
 L_\text{inner} =  \max \big\{  \|f(I_a) - f(J^f_a) \|  - \|f(I_a) - f(J^b_a) \| + 1 , 0  \big\} \end{equation}
\begin{equation}
\label{eq:triplet2}
L_\text{outer} = \max \big\{  \|f(J^f_{a}) - f(J^f_{p}) \|  - \|f(J^f_{a}) - f(J^f_{n}) \| + 1 , 0  \big\} 
\end{equation}
where $J^f_{x} = g(I_x) \odot I_x$, and $J^b_{x} = (1-g(I_x)) \odot I_x$ for $x\in{a,p,n}$. The first triplet loss $L_{\text{inner}}$  aim is to have the representation of the entire image $f(I_a)$ similar to that of the foreground $f(J^f_a)$ and dissimilar to that of the background part of the image $f(J^b_a)$, see Fig~\ref{fig:triplet}(a). The second triplet loss $L_{\text{outer}}$ is illustrated in Fig~\ref{fig:triplet}(b). The goal of this loss is to encourage the representation of masked (foreground) images of the same class to be similar, while distancing the representation of foreground images from different classes.

A  regularization term is applied to $M_{a}$ as before $\lambda R(M_a)$ to obtain a mask that is a subset of the pixels in the image.  


\noindent{\bf Architecture\quad} The encoder of $g$ is a ResNet~\cite{he2016deep}, in which the receptive field is of size $32\times 32$, i.e., the image is downscaled five times. The ResNet contains four residual blocks, with 64, 128, 256, and 512 output channels, whenever the architecture of ResNet 18 or ResNet 34 is used) and 256, 512, 1024, and 2048 output channels for the case of ResNet 50 and ResNet 101. Pre-trained weights, obtained by training on ImageNet \cite{russakovsky2015imagenet}, are used at initialization.

The decoder of $g$ consists of five upsampling blocks in order to obtain an output resolution that is identical to the original image size.  Each block contains two convolutional layers with a kernel size equal to 3 and zero padding equal to one. In addition, we use batch normalization after the last convolution layer before the activation function. The first four layers' activation function is a ReLU, while the last layer's activation function is sigmoid. Each layer receives a skip connection from the block of the encoder that has the same spatial resolution~\cite{ronneberger2015u}. 

Both our methods use the same $f$ backbone in order to obtain the localization map. This backbone is a ResNet 18~\cite{he2016deep} with four blocks with a receptive field is of size $32\times 32$. The Resnet output vector's dimension is 512, as obtained from global average pooling operator on the last feature map. This backbone is initialed with pre-trained imagenet parameters. In method I, a single Fully connected layer is used in order to obtain a prediction vector for classification. For the Siamese network, the 512 dim representation is projected linearly to a vector of 200 neurons that is used in the two triplet losses (Eq.~\ref{eq:triplet1},\ref{eq:triplet2}).

\noindent{\bf Setting the bounding box\quad} The output of $g$ is map $M$ in range 0 to 1, obtained by the sigmoid activation function. In order to derive a bounding box from this map, we follow the method of \cite{qin2019rethinking,choe2021region,choe2020evaluating}. First, a threshold $\tau$ is calculated as  
\begin{equation}
  \tau = \max(M)/10
\end{equation}
Next, we threshold all values lower than $\tau$, setting them to zero, and obtain a map $\hat M$.

We then apply the topological structural analysis method of \cite{suzuki1985topological} to the thresholded map, in order to discover the boundaries of the proposed object in the map. The method produces multiple proposals for the object contours, and we select the contour with the largest area. The bounding box is obtained by considering the bounding box of the selected contour.


\noindent{\bf Early stopping\quad}The discriminative attributes that enable the classifier $f$ to distinguish between labels that belong to visually similar categories often rely on well-localized regions of the image. As a result, the loss that we offer can lead to an over-localization of the object in the image.

In order to overcome this, we rely on an empirical observation: the pixel weighting network $g$ becomes increasingly well-localized as training progresses. In other words, $g$ becomes increasingly specific in the highlighted regions as training progresses. This way, it is able to improve the regularization term without sacrificing the loss that measures the discrepancy of the labels. 

This increase in localization is demonstrated in Fig~\ref{fig:EarlyStop}. After the first epoch, the prediction of $g$ provides a coarse mask of the object. After a few more epochs the map is much more refined and provides a good delineation of the relevant object. When training continues, the network tends to ignore non-discriminative pixels and the output becomes too partial to support segmentation of the entire object.

In order to make use of the ability of $g$ to capture the object at intermediate stages of training, while avoiding the pitfalls of using a well-trained $g$, we propose to use an early stopping procedure. Recall that the bounding box selection algorithm selects one contour. If the image has multiple contours, the bounding box would not contain these. We rely on this signal and select an epoch in which the bounding box still contains most of the weights of the mask $\hat M$.

Specifically, the epoch in which to stop is selected by considering the average score $S$ among the images in the validation set. For a single sample in the dataset, it is computed as:
\begin{equation}
  S = \frac{\sum_{x=x_1}^{x_2} \sum_{y=y_1}^{y_2} \hat M_{xy}}{ \sum_{x=0}^{H} \sum_{y=0}^{W} \hat M_{xy} }\,, 
\end{equation}
\noindent where $x_1$, $x_2$, $y_1$, $y_2$ are the bounding box coordinates for this image, and H, W are the dimensions of the image. The epoch (checkpoint) selected has the maximal average $S$ score.

\begin{table}[t]
\centering
\begin{tabular}{lccc}
\toprule
Method & GT-known & Top1 & Top1\\
 & loc[\%] & loc[\%] & cls[\%]\\
\midrule
CAM (Zhou, 2016) & 56.00 & 43.67 & 80.65\\
ACoL (Zhang, 2018) & 59.30 & 45.92 & 71.90\\
SPG (Zhang, 2018) & 58.90 & 48.90 & -\\
DANet (Xue, 2019) & 67.00 & 52.52 & 75.40\\
RCAM (Zhang, 2020) & 70.00 & 53.00 & -\\
ADL (Choe, 2019) & 75.40 & 53.04 & 80.34\\
I2C (Zhang, 2020) & 72.60 & 55.99 & 76.70\\
infoCAM+ (Qin, 2019) & 75.89 & 54.35 & 73.97\\
PsyNet (Baek, 2020) & 80.32 & 57.97 & 69.67 \\
RDAP (Choe, 2021) & 82.36 & 65.84 & 75.56\\
ART (Singh, 2020) & 82.65 & 65.22 & 77.51\\
Ours (method I) & 82.85 & 67.00 & 79.56\\
Ours (method II) & {\bf83.03}& {\bf 67.12} & 79.56\\
\bottomrule
\end{tabular}
\caption{Results on the CUB benchmark}
\label{tab:CUB object recognition}
\medskip
\centering
\begin{tabular}{lccc}
\toprule
Method & GT-known & Top1 & Top1\\
 & loc[\%] & loc[\%] & cls[\%]\\
\midrule
CAM (Zhou, 2016) & 65.2 & 56.8 & 88.9\\
HaS (Singh, 2017) & 87.4 & 76.6 & 87.6\\
ADL (Choe, 2019) & 82.8 & 73.8 & 88.9\\
RDAP (Choe, 2021) & 92.9 & 84.1 & 89.7\\
Ours (method I) & {\bf 96.1} & {\bf84.9} & 87.9\\
Ours (method II) & 95.1 & 83.7 & 87.9\\
\bottomrule
\end{tabular}
\caption{Results for the Stanford cars benchmark.}
\label{tab:CARS}
\medskip
\centering
\begin{tabular}{@{}l@{~}cc@{}}
\toprule
Method & GT-known-loc[\%] & Top1-loc[\%]\\
\midrule
CAM (Zhou, 2016) & 54.56 & 40.55\\
infoCAM (Qin, 2019) & 57.79 & 43.34\\
infoCAM+ (Qin, 2019) & 57.71 & 43.07\\
Ours (method I) & 60.21 & 43.80\\
Ours (method II) & {\bf60.41} & {\bf44.00}\\
\bottomrule
\end{tabular}
\caption{Results for Tiny-imagenet. In all methods, the classifier is a Resnet50.}
\label{tab:tiny object recognition}
 \end{table}

 \begin{table}[t]
 \begin{minipage}{.2223485\textwidth}%
            \begin{tabular}{@{}l@{~~}c@{}}
            \toprule
            Method & PxAP\\
            \midrule
            CAM \cite{zhou2016learning} & 62.57 \\
            ART \cite{singh2020attributional}  & 75.45 \\
            Ours (method I) & 76.30 \\
            Ours (method II) & {\bf 76.70} \\
            \bottomrule
            \end{tabular}
            \caption{Results for CUB \cite{wah2011caltech} segmentation. The PxAP score aggregates the average precision over multiple thresholds.}
            \label{tab:cub_seg}
            \end{minipage}\hfill%
        \begin{minipage}{.22485\textwidth}%
         \begin{tabular}{@{}l@{~~}c@{}}
        \toprule
        Method & PxAP\\
        \midrule
        CAM \cite{zhou2016learning} & 69.0 \\
        HaS \cite{singh2017hide} & 63.1 \\
        ADL\cite{choe2019attention} & 69.8 \\
        RDAP \cite{choe2021region} & 71.4 \\
        Ours (method I) & {\bf 75.6}\\
        Ours (method II) & 75.2\\
        \bottomrule
        \end{tabular}
        \caption{Results for oxford flowers segmentation.}
        \label{tab:flowers}
        \end{minipage}
\end{table}

\section{Experiments}

We use the exact same benchmark settings as the baseline methods, using the provided train/test split and evaluation protocol. The datasets supply object localization ground-truth annotation only for the validation and the test sets. The supervision signal that is provided during training is the global class-level annotation of the object. No part-level annotation is used, and we do not compare with methods that rely on such annotation. 



\noindent{\bf Benchmarks\quad} We evaluate our algorithm on the three most popular publicly available fine-grained datasets: (1) CUB-200-2011 \cite{wah2011caltech} (2) Stanford Cars \cite{krause20133d} (3) Oxford-flowers \cite{nilsback2008automated}. In addition, we test our proposed algorithm on the generic image classification dataset  tiny-imagenet \cite{le2015tiny}. 

CUB-200-2011 \cite{wah2011caltech} contains 200 birds species, with 11,788 images divided into 5994 training images and 5794 test images. The images were taken at different resolutions and aspect ratios where the object is not centralized. The  variations in the dataset, such as pose, viewpoint and illumination increase the complexity of the task. 


Stanford Car \cite{krause20133d} has 196
categories of cars, which differ in at least one of three attributes: manufacturer, model, and year. There are 8144 samples in the training set and 8041 samples in the test set.  

Oxford-102 \cite{nilsback2008automated} contains 102 categories of flowers, with 1020 training images and 6149 test images. The dataset supplies mask annotations (not just bounding boxes), which enables the utilization of segmentation metrics.

Tiny-ImageNet is a small version of ImageNet, with a reduced number of classes, number of instances per class, and image resolution (64*64). It consists of 200 classes, with 500 training images and 50 validation images per class. Unlike CUB-200-2011 and other fine-grained classification datasets, Tiny-ImageNet contains various object types. Training classifiers on Tiny-ImageNet is faster than training on imagenet, due to the smaller number of samples and their reduced dimensions. However, for the same reasons, obtaining high classification accuracy is also challenging.

\begin{figure*}[t]
    \setlength{\tabcolsep}{4.5pt} 
    \renewcommand{\arraystretch}{2} 
    \begin{tabular}{cccccccc}
    
    \includegraphics[width=0.105\linewidth]{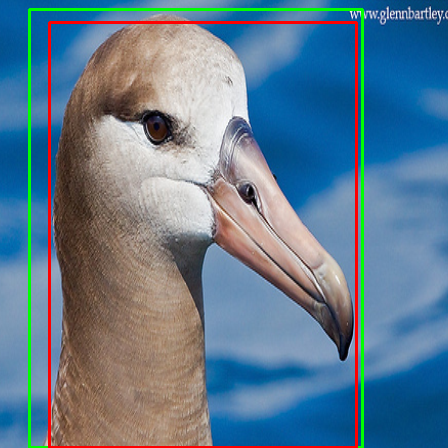} &
    \includegraphics[width=0.105\linewidth]{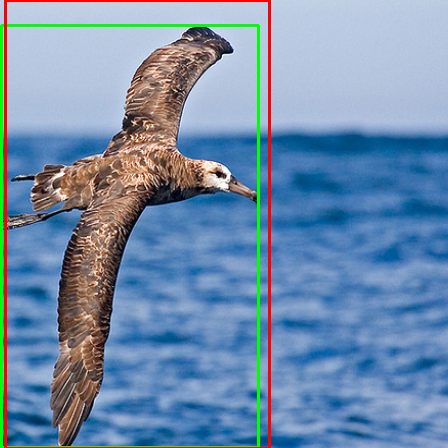} &
    \includegraphics[width=0.105\linewidth]{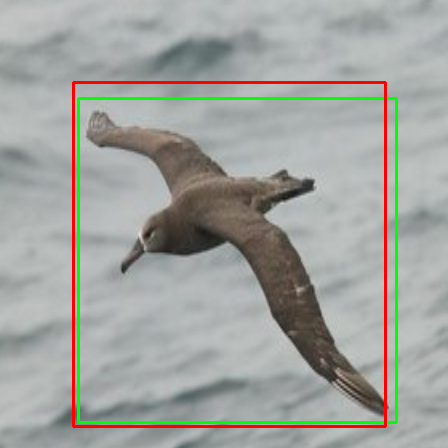} &
    \includegraphics[width=0.105\linewidth]{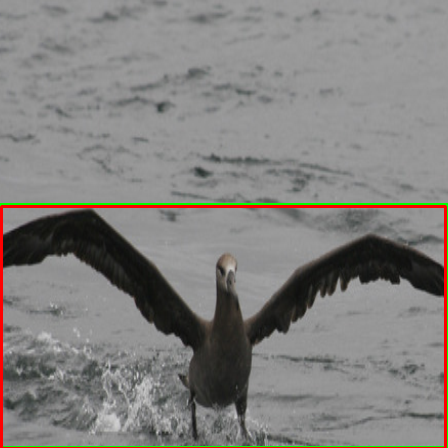} &
    \includegraphics[width=0.105\linewidth]{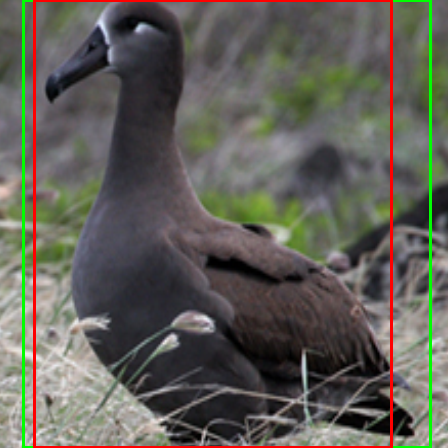} &
    \includegraphics[width=0.105\linewidth]{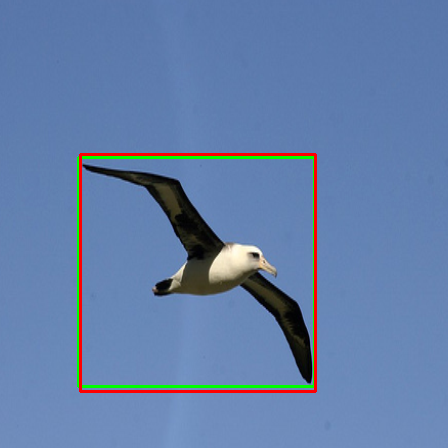} &
    \includegraphics[width=0.105\linewidth]{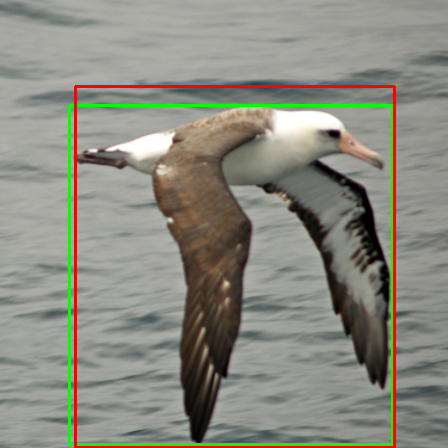} &
    \includegraphics[width=0.105\linewidth]{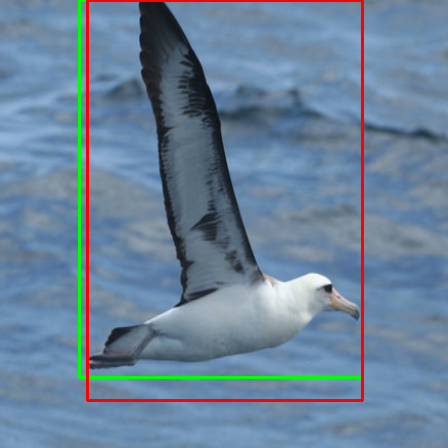} \\
    
    \includegraphics[width=0.105\linewidth]{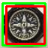} &
    \includegraphics[width=0.105\linewidth]{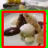} &
    \includegraphics[width=0.105\linewidth]{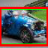} &
    \includegraphics[width=0.105\linewidth]{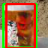} &
    \includegraphics[width=0.105\linewidth]{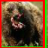} &
    \includegraphics[width=0.105\linewidth]{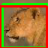} &
    \includegraphics[width=0.105\linewidth]{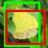} &
    \includegraphics[width=0.105\linewidth]{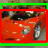} \\

    \includegraphics[width=0.105\linewidth]{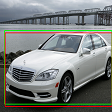} &
    \includegraphics[width=0.105\linewidth]{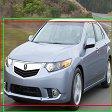} &
    \includegraphics[width=0.105\linewidth]{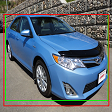} &
    \includegraphics[width=0.105\linewidth]{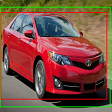} &
    \includegraphics[width=0.105\linewidth]{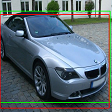} &
    \includegraphics[width=0.105\linewidth]{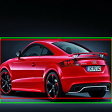} &
    \includegraphics[width=0.105\linewidth]{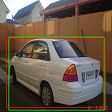} &
    \includegraphics[width=0.105\linewidth]{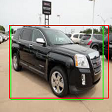} \\
    
    \end{tabular}
    \caption{Samples results from CUB, CAR, and Tiny ImageNet. The green bounding boxes represent the ground truth and the red ones represent our predicted bounding box. For brevity, all visual results in this paper are obtained with method I. See supplementary for the (quite similar) results of method II. } 
    \label{fig:samples results}
\end{figure*}

\begin{figure}[t]
    \setlength{\tabcolsep}{4.5pt} 
    \renewcommand{\arraystretch}{2} 
    \begin{tabular}{ccc}

    \includegraphics[width=0.285\linewidth]{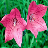} &
    \includegraphics[width=0.285\linewidth]{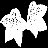} &
    \includegraphics[width=0.285\linewidth]{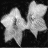} \\

    \includegraphics[width=0.285\linewidth]{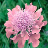} &
    \includegraphics[width=0.285\linewidth]{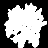} &
    \includegraphics[width=0.285\linewidth]{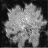} \\

    (a) & (b) & (c)\\
    \end{tabular}
    \caption{Oxford Flowers 102 results where (a) is the input image (b) is the ground-truth (c) output map $M$.} 
    \label{fig:flowers}
\end{figure}

\begin{figure}[t]
    \centering
    \begin{tabular}{cccc}
    
    \includegraphics[width=0.21\linewidth]{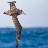} &
    \includegraphics[width=0.21\linewidth]{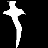} &
    \includegraphics[width=0.21\linewidth]{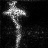} &
    \includegraphics[width=0.21\linewidth]{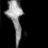} \\

    \includegraphics[width=0.21\linewidth]{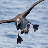} &
    \includegraphics[width=0.21\linewidth]{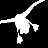} &
    \includegraphics[width=0.21\linewidth]{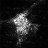} &
    \includegraphics[width=0.21\linewidth]{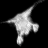} \\
    \end{tabular}
    \caption{CUB weakly supervised segmentation where the first row is the input image, the second row is the ground-truth mask, the third row is the output map of \cite{singh2020attributional} and the last row is our output map $M$.} 
    \label{fig:cub_seg}
\end{figure}

On all datasets, except for Oxford-102, we compute two accuracy scores. For both scores, an intersection over union above 0.5 indicates correct localization. In the GT-known-loc score, we compare the bounding box even if the classifier was mistaken. In the Top1-loc, we only consider the result to be correct if the classifier has predicted the correct class for the image. While we claim no contribution to the classifier, we present this score as well, since it is prevalent in the literature. Moreover, we note that we outperform other methods by this score, even though the top-1 classification of our method, reported as Top1-cls is not as high as that of other methods. 

For the CUB and Oxford datasets, which contain foreground-background masks, we also compute a segmentation metric. Specifically, we use the pixel average precision (PxAP) metric proposed by \cite{choe2020evaluating}. It relies on the following definitions:

\begin{equation}
PxPrec(\sigma) = \frac{ \mid  \big\{ {M_{ij}^{(n)} \geq \sigma} \big\}  \cap \big\{  T_{ij}^{(n)} = 1\big\}\mid }{\mid  \big\{ M_{ij}^{(n)} \geq \sigma \big\}\mid }
\end{equation}
\begin{equation}
PxRec(\sigma) = \frac{ \mid  \big\{ {M_{ij}^{(n)} \geq \sigma} \big\}  \cap \big\{  T_{ij}^{(n)} = 1\big\}\mid }{\mid  \big\{  T_{ij}^{(n)} = 1\big\}\mid }
\end{equation}
\noindent where $i,j$ are the indices of the pixels, $M$ is the weighting map, and $T$ is the ground-truth mask.

PxAP is the area under the precision-recall curve
\begin{equation}
\nonumber
PxAP : =  \sum_l PxPrec(\sigma_l)(PxRec(\sigma_l) - PxPrec(\sigma_{l-1}))
\end{equation}


\noindent{\bf Implementation details\quad} While the method has many hyperparameters, listed below, all of these are fairly standard and no attempt was done to tune them in any way. The test data was not used in any way to determine the hyperparameters or the early stopping epoch.

In our method, the input image is first resized to 256$\times$256, and then a crop is applied to reduce it to size 224$\times$224. During training, cropping is applied at a random image location, while during validation and test, we take the crop from the center of the image. 

The classifier $f$ based on Resnet18 backbone for fine-grained datasets, and Resnet50 for tiny-imagenet. The optimizer is SGD with a batch size of 16, and an initial learning rate of 0.001 for 200 epochs, the weight decay is 1e-4. The scheduler decreases the learning rate by a factor of 10 every 80 epochs. For augmentation, the training procedure employs a resize to a fixed size followed by a random crop, as well as a random horizontal flip. During inference, the algorithm employs resize and central crop. The classifier consists of a single fully-connected layer $R^{k} \to R^{d}$, where $d$ is a number of classes, and k is the latent space size. 

Network $g$ for the fine-grained recognition datasets is trained with the SGD optimizer with a batch size of 128, and an initial learning rate of 0.0001 for 100 epochs. The tiny imagenet model is trained with the same SGD optimizer, a batch size of 128, weight decay 5e-5, an initial learning rate of 0.0001, and for 1000 epochs. 

During the phase of training both $f$ and $g$, random horizontal flip with 0.5 probability is applied as augmentation. Training of all models takes place on a single GeForce RTX 2080Ti Nvidia GPU. This attests to the efficiency of our methods but prevents us from running on much larger generic classification datasets.

\noindent{\bf Results\quad} Tab.~\ref{tab:CUB object recognition} compares our results to the existing  methods for CUB-200-2011. The accuracy of method I is 82.85\%  for ground-truth known localization and 67\% for top-1 localization accuracy, which outperforms all other methods. Method II is slightly better on this dataset. Note that the top-1 localization accuracy is high even though our classifier is standard and not the leading one (we make no claims regarding the performance of the classifier).

On the Stanford Car \cite{krause20133d} we obtain the baseline results from \cite{choe2021region}, who ran all methods cleanly. The results presented in Tab.~\ref{tab:CARS} indicate that both our methods outperforms all baselines in the GT-known localization by a larger margin than that obtained between the previous work. There is an advantage to method I. In the Top1 loc score, which also incorporates the classifier accuracy, the simplicity of our accuracy eliminates some of the gap from the best baseline method (RDAP). Method I is still best. However, Method II is outperformed by this method.

Tab.~\ref{tab:tiny object recognition} summarises the results for the Tiny-imagenet dataset. All baseline methods employ the same ResNet50 classifier.  Both our methods outperform the baseline by a margin that is larger than the difference between the baseline methods, with method II showing a slight advantage.

The output of our method, in comparison to the ground truth, is presented in Fig~\ref{fig:samples results}. As can be seen, our method can match the ground truth bounding box well and is not overly focused on small discriminative regions.

The results that evaluate the weight map obtained from $g$ as a segmentation map are presented in Tab.\ref{tab:flowers} and Tab.\ref{tab:cub_seg} for the Oxford flowers and CUB datasets, respectively. Evidently, the PxAP score for both our methods is considerably higher than for the other methods, by a sizable margin.  On Oxford the first of our methods shows an  advantage, and on CUB there is a preference for method II over method I. 

\begin{figure}
    \centering
    \begin{tabular}{cccc}
    \includegraphics[width=0.17\linewidth]{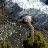} &
    \includegraphics[width=0.17\linewidth]{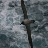} &
    \includegraphics[width=0.17\linewidth]{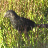} &
    \includegraphics[width=0.17\linewidth]{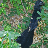} \\

    \includegraphics[width=0.17\linewidth]{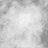} &
    \includegraphics[width=0.17\linewidth]{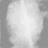} &
    \includegraphics[width=0.17\linewidth]{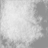} &
    \includegraphics[width=0.17\linewidth]{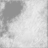} \\

    \includegraphics[width=0.17\linewidth]{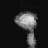} &
    \includegraphics[width=0.17\linewidth]{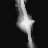} &
    \includegraphics[width=0.17\linewidth]{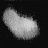} &
    \includegraphics[width=0.17\linewidth]{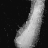} \\
    \end{tabular}
    \caption{An ablation experiment. Top: input image, middle: the results obtained without the regularization term in Eq.~\ref{eq:one}, bottom: the results of our method.} 
    \label{fig:no_reg}
\end{figure}

Sample results on the Oxford-flowers102 dataset are presented in Fig.~\ref{fig:flowers}. As can be seen, the output of $g$ matches the ground-truth masks well. Fig.~\ref{fig:cub_seg} presents the output masks of our algorithm and of ART~\cite{singh2020attributional} for samples from the CUB-200-2011 dataset. As can be seen, our output is considerably more uniform than that of ART even though the numerical evaluation indicates only a modest gap in performance for this specific dataset. 

As a first ablation experiment, we remove the regularization parameter in Eq.~\ref{eq:one}. Without this term, a uniform $g$ that outputs 1 everywhere would reach a zero loss. In practice, as can be seen in Fig.~\ref{fig:no_reg} the network converges to another low-loss solution that covers most of the image. 

As for method II, Tab.~\ref{tab:cub_ablation} summaries the impact of each term in the loss function on CUB WSOL dataset. Both of the triplet losses are impact the performances and essential in order to obtain optimal results. The importance of the regularization is also evident in this ablation.

\begin{table}
\centering
\begin{tabular}{cccc}
\toprule
$R(g(I))$ & $L_{outer}$ & $L_{inner}$ & GT-known-loc[\%]\\
\midrule
- & $\surd$ & $\surd$ & 62.13 \\
$\surd$ & - & $\surd$ & 81.64 \\
$\surd$ & $\surd$ & - & 79.92 \\
$\surd$ & $\surd$ & $\surd$ & 83.03 \\
\bottomrule
\end{tabular}
\caption{Ablation results for method II on CUB \cite{wah2011caltech} WSOL.}
\label{tab:cub_ablation}
\end{table}

\section{Discussion}

Many of the recent WSOL methods improve the underlying classifier $f$ in order to better match the needs of localization tasks. For example, \cite{singh2017hide} covers the discriminative parts during the training of the classifier in order to have it rely on additional regions. For a similar reason, \cite{zhang2018adversarial} train two classifiers, where one covers the feature maps of the other. \cite{zhang2020rethinking} improves the representation of the classifier by employing quantization. In contrast, our method employs a standard pre-trained classifier and obtains state-of-the-art results out-of-the-box.

Furthermore, we note that $f$ does not need to be a classifier. We plan, as future work, to use self-supervised representations such as SwAV \cite{caron2020unsupervised} as well as text-image matching neural networks such as CLIP~\cite{radford2021learning}. 

\section{Conclusions}

We present a new WSOL method that relies on a novel and elegant training loss that produces leading segmentation results, and with minimal post-processing also state-of-the-art localization. Unlike methods that rely on explainability, the classifier $f$ serves as a black-box and only a few assumptions are made regarding its nature.

Over-segmentation is prevented by applying an early stopping criterion that does not require any pixel-level labels, and without modifying the training of $f$ in any way. The method is simple to implement and optimize and our code is attached as supplementary.



\section*{Acknowledgment}
This project has received funding from the European Research Council (ERC) under the European
Unions Horizon 2020 research and innovation programme (grant ERC CoG 725974).

{\small
\bibliographystyle{ieee_fullname}
\bibliography{egbib,explainability,selfsupervised}
}

\end{document}